%% file: colm2026_conference.tex
\documentclass{article}
\usepackage[final]{colm2026_conference}

\usepackage{microtype}
\usepackage{hyperref}
\usepackage{url}
\usepackage{booktabs}
\usepackage{multirow}
\usepackage{graphicx}
\usepackage{amsmath}
\usepackage{amssymb}
\usepackage{subcaption}
\usepackage{array}
\usepackage{lineno}
\usepackage{wrapfig}
\usepackage{algorithm}
\usepackage{algpseudocode}
\usepackage{cleveref}

\usepackage{xspace}

\definecolor{darkblue}{rgb}{0, 0, 0.5}
\hypersetup{colorlinks=true, citecolor=darkblue, linkcolor=darkblue, urlcolor=darkblue}

\title{
More Than Words: Compositional Tokenization for Efficient Language Models
}
\author{Yuval Reif \;\;\;\;\; Guy Kaplan \;\;\;\;\; Roy Schwartz\\
  The Hebrew University of Jerusalem \\
  \texttt{\normalsize \{yuval.reif,guy.kaplan3,roy.schwartz1\}@mail.huji.ac.il}}

\newcommand{\methodbpe}{\textsc{CoBPE}\xspace}
\newcommand{\methodtoken}{compound token\xspace}

\newcommand{\methodtokens}{compound tokens\xspace}

\newcommand{\spaceprefixed}[1]{{\underline{\hspace{0.25cm}}}{#1}}

\newcommand{\nanochat}{\textsc{nanochat}\xspace}

\newcommand{\tabref}[1]{Table~\ref{tab:#1}}
\newcommand{\figref}[1]{Figure~\ref{fig:#1}}
\newcommand{\appref}[1]{Appendix~\ref{app:#1}}
\newcommand{\secref}[1]{§\ref{sec:#1}}
\newcommand{\subsecref}[1]{§\ref{subsec:#1}}

\begin{document}

\ifcolmsubmission
\linenumbers
\fi

\maketitle

\input{text/0.abstract}
\input{text/1.introduction}
\input{text/2.tokenization_motivation}

\input{text/3.method}

\input{text/4.experiments}

\input{text/5.multilingual_potential}

\input{text/999.related_work}

\input{text/999.conclusion}

\input{text/999.acknowledgments}

\bibliographystyle{colm2026_conference}
\bibliography{custom}

\clearpage
\appendix
\input{text/999.appendix}

\end{document}

%% file: text/0.abstract.tex
\begin{abstract}
Language models process and generate text sequentially in \emph{token} units, and the tokenizer determines how much text each inference step covers. Under standard tokenization, a short English phrase such as ``On the table.'' is usually produced as four separate predictions for the preposition~(\textit{On}), article~(\textit{the}), noun~(\textit{table}), and punctuation~(\textit{.}), where each consumes a sequence position and adds inference cost. We introduce \methodbpe, a compositional tokenization approach that represents such phrases as a lexical base token~(\textit{table}) attached with a small set of reusable surface modifiers, composed in embedding space at input and predicted jointly at output. In controlled pretraining from scratch at 780M and 1.3B scales, \methodbpe shortens sequences by 30\% and improves average downstream performance by 1.2 points relative to standard BPE under matched training compute.
Our results suggest that part of what is now expressed through token sequences can instead be modeled through structured representations, opening a broad design space for more token-efficient and capable language models.\footnote{Code and model weights are available at \url{https://co-bpe.github.io/}.}
\end{abstract}

%% file: text/1.introduction.tex
\section{Introduction}
\label{sec:introduction}

Large language models~(LLMs) do not operate directly on raw text, but instead consume \emph{token} sequences, most commonly produced by Byte-Pair Encoding~\citep[BPE;][]{bpe2016sennrich}. To build the model's token vocabulary, such methods first split text into whitespace- and punctuation-bounded chunks in \emph{pre-tokenization}, and then allocate tokens to frequent byte sequences within chunks~\citep{openai2024tiktoken}.

While effective, this process results in units that inherit pre-tokenization boundaries and follow surface string structure, creating two related inefficiencies. First, very common sequences such as ``\textsl{on the}'' cannot become a single token~\citep{superbpe2025}. Instead, the model must process them using multiple sequential steps, with items that carry different amounts of content such as ``\textsl{the}'', ``\textsl{table}'', and ``\textsl{,}'' all requiring the same amount of compute. Second, the same interface treats near-duplicate forms such as \texttt{table}, capitalized \texttt{Table}, and space-prefixed \texttt{\spaceprefixed{table}} as separate vocabulary entries. As a result, related tokens are represented using inconsistent symbols and require the model to learn their relation from data~\citep{schafer-etal-2024-effect,kaplan2025from}, while also consuming token slots in the size-limited vocabulary that could otherwise support other languages or domains~\citep{vocab_diet}. These choices can materially affect model efficiency, as training and inference costs depend directly on the number of tokens used to represent text~\citep{ahia2023languages,tokenizer-choice-2024}. Overall, this suggests there is still substantial room to improve compression beyond standard sub-word tokenization.

In this paper, we introduce a \textbf{Co}mpositional \textbf{BPE} tokenization approach~(\methodbpe). \methodbpe segments text into \emph{\methodtokens}, each composed of a \emph{base token} and a small set of \emph{surface modifiers}. The base token serves as the lexical anchor, while the modifiers encode leading space, capitalization, prepositions, determiners, and surrounding punctuation~(see \figref{overview}, top). On the input side, the model composes the unit in embedding space by adding the active modifier embeddings to the base embedding; on the output side, it predicts the base token and modifier labels as a structured output, with modifier predictions conditioned on the base token. Thus, phrases like ``On the table?'' can be generated as a single \methodtoken centered on \texttt{table}, rather than four separate token predictions.

\input{figures/overview}

We first quantify how much \methodbpe shortens sequences relative to standard BPE by folding frequent markers such as determiners, prepositions, and punctuation into the same structured unit as a base token~(\figref{overview}, bottom). We then describe how a standard decoder-only transformer can be adapted to consume and generate \methodtokens through compositional input embeddings and factorized output prediction.

Next, we study \methodbpe through controlled English pretraining from scratch on the ClimbMix corpus~\citep{nemotronclimb2025} at 780M and 1.3B scales. Across the main experiments, we hold the model family, architecture, training recipe, vocabulary budget, and evaluation suite fixed and vary only the tokenization scheme.

In experiments, \methodbpe improves average downstream performance over standard BPE under matched training-compute budgets while also shortening sequences substantially, allowing more text to be processed under the same budget. At both model scales, it yields stronger downstream performance on the same 30-task evaluation suite, improving the average score by 1.2 points while reducing sequence length by 30\%. To test whether these gains translate into end-to-end efficiency, we evaluate \methodbpe on the \nanochat speedrun benchmark~\citep{nanochat}, which measures how quickly a model reaches a fixed capability target. In that setting, \methodbpe reduces training time by 18.6\%, showing that its compression gains translate into faster end-to-end training.

Taken together, our results reveal a broader design space for language modeling, in which some of the structure now carried by sequence length can instead be modeled directly within the token representation itself.
In \methodbpe, this takes the form of a shared lexical base combined with a small set of reusable modifiers, yielding a tokenizer that is both more compact and more effective under standard language-model training. More broadly, it points to structured token representations as a promising alternative to standard tokenization.

%% file: figures/overview.tex
\begin{figure}[t]
    \centering
    \begin{subfigure}{0.98\linewidth}
        \centering
        \includegraphics[width=\linewidth]{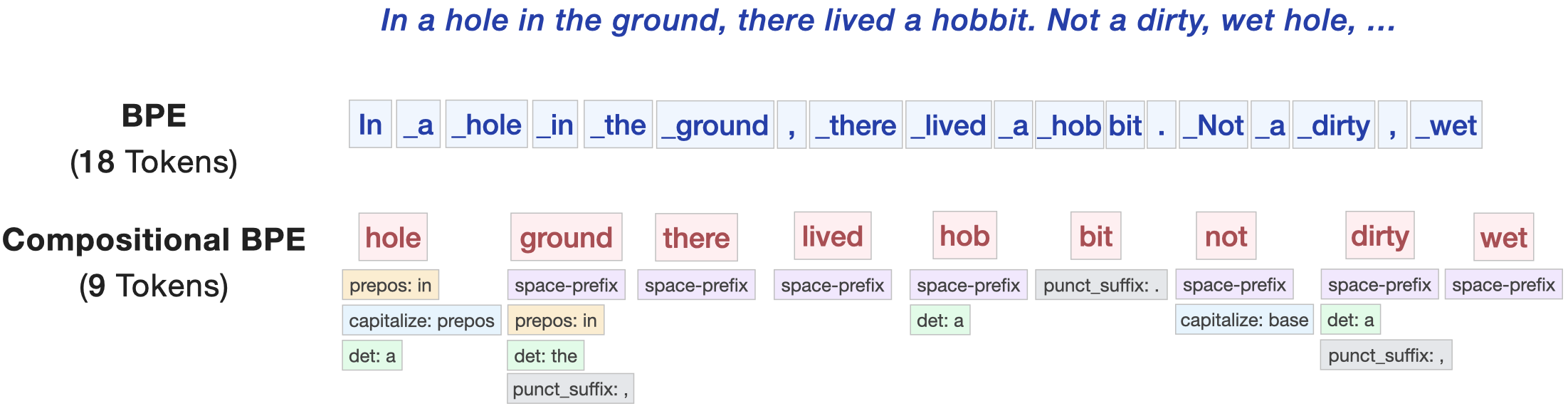}
    \end{subfigure}

    \vspace{0.5em}

    \begin{subfigure}{0.7\linewidth}
        \centering
        \includegraphics[width=\linewidth]{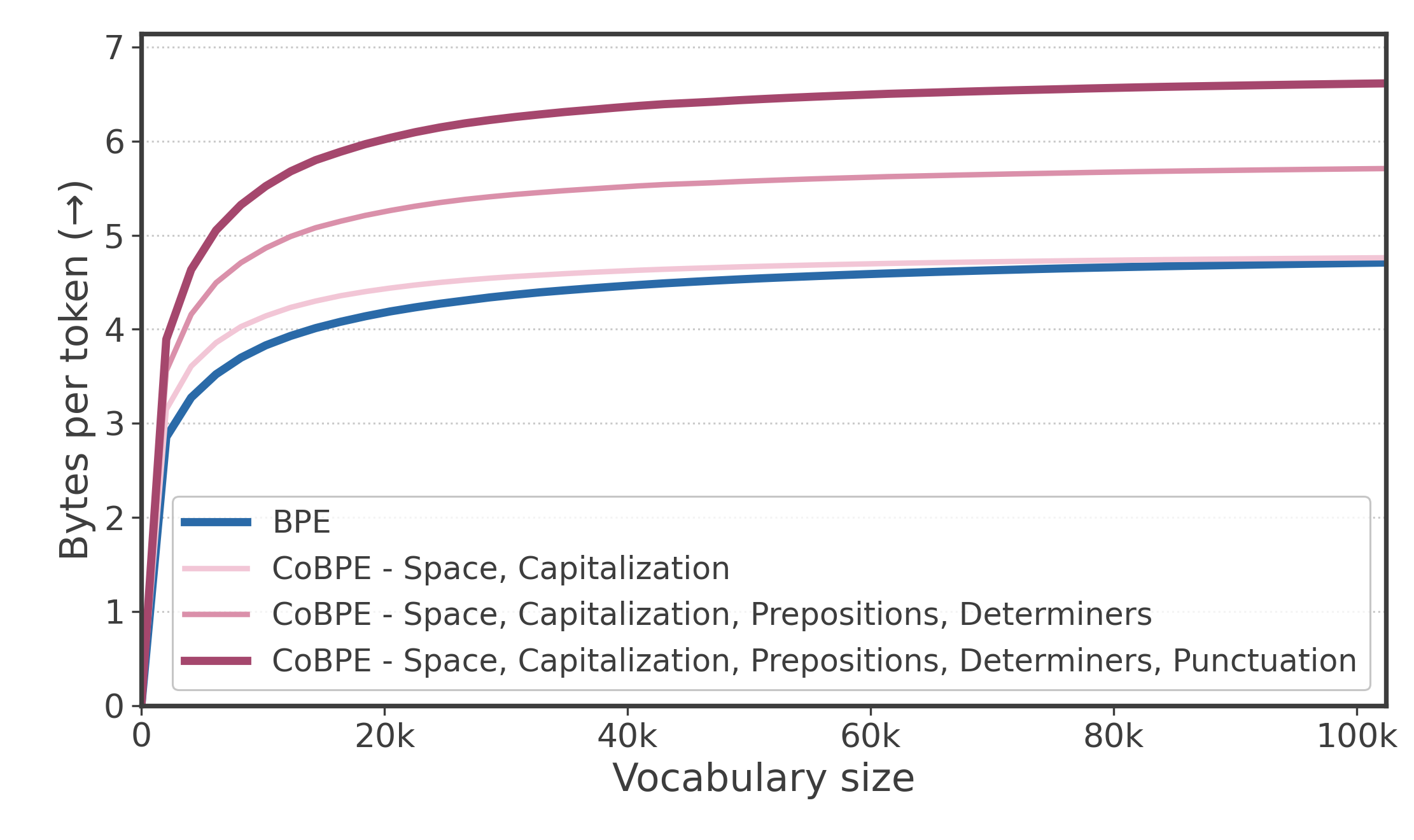}
    \end{subfigure}
    \caption{\textbf{Compositional tokenization shortens sequences compared to BPE by factoring recurring surface markers around a lexical base.} \textbf{Top:} standard BPE represents the excerpt as a sequence of 18 tokens, while Compositional BPE~(\methodbpe) represents the same text using only 9 \textsl{\methodtokens} built from a base token together with attached modifiers for leading space, capitalization, determiners, prepositions, and surrounding punctuation. \textbf{Bottom:}
    Encoding efficiency (bytes-per-token) of English text for BPE and \methodbpe as a function of vocabulary size. Adding modifier families progressively improves compression over BPE, and the full \methodbpe configuration covers roughly 1.4$\times$ more bytes per token across vocabulary sizes.
    }
    \label{fig:overview}
\end{figure}

%% file: text/2.tokenization_motivation.tex
\section{Tokenization beyond flat sequences}
\label{sec:compression-motivation}

In modern LLMs, tokenization determines how much text each model step represents. Because every sequence position incurs computation, this choice directly affects training and inference costs~\citep{ahia2023languages,petrov2023language}.
The standard approach is BPE tokenization with word-boundary pre-tokenization~\citep{bpe2016sennrich, dagan2024getting}, which greedily builds a token vocabulary of a target size before pretraining. Given a large text sample, the text is first split into word-bounded chunks in pre-tokenization, and the vocabulary is then built by repeatedly merging the most frequent adjacent character or sub-word sequences within those chunks until the target size is reached, allocating tokens to the results of each merge.

At sufficiently large vocabulary sizes, BPE is highly effective for compression: in English, modern tokenizers use roughly 1.2 tokens per word on average~\citep{tokenizer-choice-2024}, with most words represented by a single token, each receiving its own inference step. However, this design allocates similar compute to process \textsl{function words}---frequent words expressing grammatical relations (\emph{the}, \emph{of}, \emph{on})---as it does to \textsl{content words}, which carry most of the lexical meaning (such as nouns, main verbs and adjectives). Punctuation is treated similarly: with pre-tokenization, punctuation marks are typically separate tokens rather than part of surrounding words. Importantly, this suggests that even after BPE's strong compression, considerable compute is spent on processing standalone grammatical tokens.

Recent work relaxes word-boundary constraints in pre-tokenization and allows multi-word tokens to form (e.g., ``\emph{in the ground.}''), substantially improving compression and downstream efficiency~\citep{superbpe2025,schmidt2025boundless}. This compression, however, is phrase-specific: for example, instead of representing ``\emph{the}'' as a reusable unit across contexts, the vocabulary must allocate separate entries to combinations such as ``\emph{the ground}'', ``\emph{the earth}'', and similar phrases. Pushing this compression strategy further requires adding increasingly specific entries such as ``\emph{in the ground?}'' or ``\emph{On the ground!}'', making it impractical to capture the richness of language even under large vocabulary sizes.

\input{figures/grammar_marker_share_wrap}
These observations suggest that attaching such markers to neighboring words could substantially improve compression.  To measure this opportunity, we quantify the fraction of token positions occupied by grammatical markers in English FineWeb documents~\citep{penedo2024fineweb}. We analyze text tokenized with GPT-4's tokenizer~\citep{openai2024tiktoken}, which several modern LLMs adopt as a starting point for their own tokenizers~\citep{qwen2.5,llama3}. We measure the share of positions taken by standalone determiners~(e.g., \emph{the}, \emph{a}, \emph{this}), prepositions~(\emph{by}, \emph{in}, etc.), and punctuation.
In \figref{grammar-marker-share}, we show that these categories account for 29.8\% of all token positions,\footnote{See \figref{grammar-marker-detail} in \appref{appendix-analysis} for a detailed view of the most frequent markers within each category.} despite carrying relatively little lexical content~\citep{jiang-etal-2023-llmlingua,Bitton:2021}.
This indicates that much of the remaining compression potential for natural language comes not from encoding rarer words, but from explicitly modeling frequent grammatical and orthographic structure surrounding words.

%% file: figures/grammar_marker_share_wrap.tex
\begin{wrapfigure}{r}{0.45\linewidth}
    \centering
    \includegraphics[width=\linewidth]{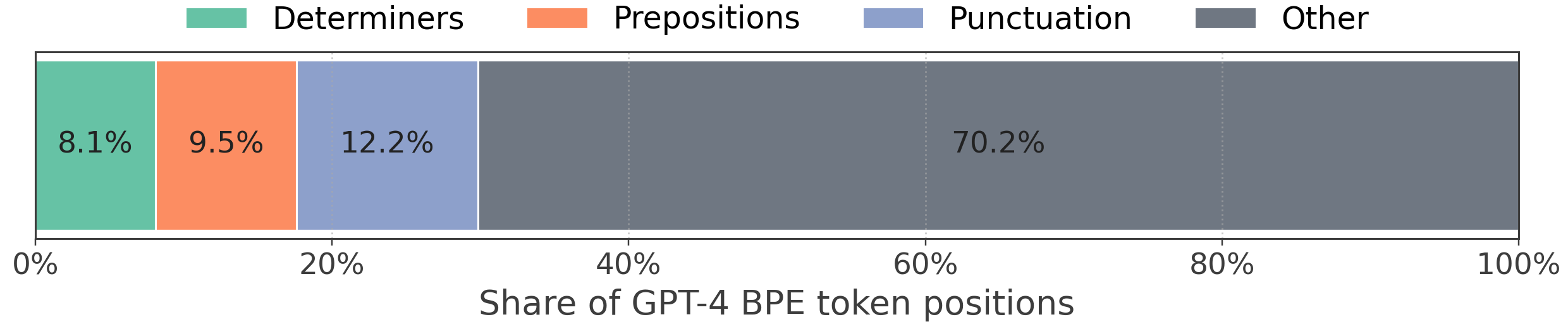}
    \caption{\textbf{A substantial share of token positions under a modern BPE tokenizer are grammatical markers.} Breakdown of token positions occupied by determiners, prepositions, and punctuation in FineWeb text tokenized with GPT-4.}
    \label{fig:grammar-marker-share}
\end{wrapfigure}

%% file: text/3.method.tex
\section{\methodbpe: compositional BPE tokenization}
\label{sec:method}

The potential to compress frequent grammatical markers~(\secref{compression-motivation}) motivates a simple reformulation of standard BPE: rather than dedicating separate prediction steps or vocabulary entries to frequent grammatical material or surface variation, we can represent these as \textsl{modifiers} attached to lexical \textsl{base tokens} and process each composition in a single inference step.

We call this approach Compositional BPE tokenization (\methodbpe). We next define the tokenizer and its modifier inventory~(\subsecref{representation}), show how it improves compression relative to BPE~(\subsecref{method_compression}), and describe the adaptation needed for a language model to consume and generate \methodtokens~(\subsecref{modeling}).

\subsection{Definition}
\label{subsec:representation}

Standard BPE segments a phrase like ``\emph{in the ground,}'' into four tokens: preposition, determiner, noun, and punctuation.
Under \methodbpe, such spans are represented as a single structured unit we call a \textsl{\methodtoken}. Each \methodtoken is built around a lexical \emph{base token}, along with a small set of attached \emph{modifiers} encoding frequent markers such as determiners, prepositions, and punctuation. Thus, the span above is represented as the base token \texttt{ground} along with the modifiers \texttt{preposition=in}, \texttt{determiner=the}, and \texttt{punct-suffix=,}.

BPE also treats common surface variants as distinct token strings. For example, \texttt{ground}, \texttt{Ground}, and \texttt{\_ground} (with a leading space, as in mid-sentence occurrences) are separate tokens despite sharing the same lexical core. \methodbpe factors out this variation by keeping the base token fixed and representing leading space and capitalization as dedicated modifiers.

\Cref{tab:tokenization-examples} shows the \methodbpe decomposition for several related spans.
We focus on English and define eight modifier groups:
four categorical groups for determiners, prepositions, prefix punctuation, and suffix punctuation; three binary indicators for whether the base token, determiner, and preposition are capitalized; and one binary indicator for the presence of a leading space before the span. We focus on markers that can be identified and reconstructed using local surface rules, with exact values and exclusions listed in~\appref{tokenizer-details}.\footnote{Supporting other types of modifiers~(e.g., morphological modifiers, \citealp{vocab_diet}) requires external resources and is thus deferred to future work.}

Formally, \methodbpe maps a text $t$ into a sequence of \methodtokens $c_1, c_2, \ldots, c_N$, where
\[
c_i = \left(b_i, m_i^{(1)}, \ldots, m_i^{(G)}\right)
\]
Here, \(b_i\) is a lexical base token, and each \(m_i^{(g)}\) is the value assigned to one modifier group. Together, the base token and modifier values specify the surface form realized at position \(i\).
If a word spans multiple tokens, \methodbpe distributes modifiers across it deterministically~(see \texttt{[hob, bit]} in \figref{overview}): leading-space, capitalization, determiner, preposition, and prefix-punctuation markers are attached to the first token in the span, while suffix punctuation is attached to the last token.

Having defined the representation, we next describe how the tokenizer constructs it.
To construct a \methodbpe tokenizer, we first build a BPE tokenizer with a pre-tokenizer that separates punctuation and leading spaces from word chunks. To avoid allocating separate vocabulary entries to capitalized words that \methodbpe later represents through lowercase bases, we train this tokenizer on text in which standard titlecased words are lowercased.\footnote{For simplicity, we restrict capitalization modifiers to standard titlecase. ALL CAPS and camelCase are left unchanged and handled by the underlying BPE tokenizer, and broader capitalization handling is deferred to future work.} Function words and punctuation markers are still included in this underlying BPE vocabulary; they may later be attached as modifiers when composition is possible, but they also remain available as ordinary tokens when no valid attachment applies.

At inference time, \methodbpe processes text in two stages.
The text is first tokenized with the underlying BPE tokenizer, which produces an ordinary BPE token stream containing lexical items, determiners, prepositions, and punctuation as separate tokens. A deterministic left-to-right pass then identifies eligible base tokens.\footnote{An eligible base token is any non-punctuation/numeric token that does not occur mid-word.}
For each base, we attach neighboring determiner, preposition, and punctuation markers from a closed inventory of modifiers, along with capitalization and leading-space cues.\footnote{This attachment rule is purely local and does not require syntactic parsing. For example, in ``by the large house'', the markers ``by the'' attach to ``large'' rather than the syntactic head ``house''. Importantly, no information is discarded: every component remains represented in the sequence, allowing the model to learn the relevant relationships from context, as in standard tokenization. We defer syntax-aware attachment to future work.}
Tokens that cannot be validly attached remain as ordinary positions in the final sequence.
The resulting sequence is the \methodbpe representation used throughout training and evaluation.
See \appref{tokenizer-details} for tokenizer construction details, modifier inventories and usage statistics, and pseudocode for the inference-time composition pass.

As we show next, \methodbpe allows the tokenizer to represent text in fewer sequence positions while preserving the surface information needed to reconstruct the original text.

\input{tables/tokenization_examples}

\subsection{Compression relative to BPE}
\label{subsec:method_compression}

\methodbpe improves compression through two complementary mechanisms.
First, function-word and punctuation modifiers shorten sequences by absorbing frequent grammatical material into a single prediction.
Second, modifiers for leading space and capitalization let related surface forms share a single lexical token~\citep{vocab_diet}, freeing vocabulary capacity from near-duplicate tokens and allowing additional unique tokens to be allocated.

We measure compression with \emph{bytes per token} (BPT), the average number of UTF-8 bytes covered by each token on a large corpus; higher BPT means that the same text is represented using fewer tokens. We construct BPE and \methodbpe tokenizers on ClimbMix~\citep{nemotronclimb2025} and evaluate BPT on FineWeb~\citep{penedo2024fineweb}. For each tokenizer, we plot the BPT as a function of vocabulary size. To isolate the contribution of different modifier types, we also build \methodbpe variants incrementally: first adding leading-space and capitalization modifiers, then determiners and prepositions, and finally punctuation.

We report the results in \figref{overview}~(bottom). Sharing entries across capitalization and leading-space variants already improves BPT at smaller vocabulary sizes, indicating that BPE spends much of its vocabulary on near-duplicate forms.
For example, in a 32k BPE vocabulary, 15\% of entries correspond to capitalized variants and 12\% to space-prefixed forms; under \methodbpe, these are represented through modifiers, allowing new, unique tokens to be allocated at the same vocabulary budget.
Adding determiners, prepositions, and punctuation leads to large, consistent improvements across all vocabulary sizes, as these markers occur frequently enough to collapse many multi-token spans into a single \methodtoken.
Overall, \methodbpe improves BPT by roughly 40\% across vocabulary sizes.
This suggests that much of the remaining compression potential beyond standard BPE lies in modeling local structure around words, not only in better segmenting the words themselves.

\subsection{Adapting language models to \methodbpe}
\label{subsec:modeling}

\input{figures/method_illustration}
Given a piece of text tokenized by \methodbpe, a standard decoder-only transformer can be adapted to operate over \methodtokens by modifying the input and output parameterizations.
The transformer still processes one sequence position at a time, but each position now corresponds to a structured unit rather than a single surface token.

On the input side, each \methodtoken is represented by composing the embedding of its lexical base with embeddings for its active modifiers. Let $E_{\text{base}} \in \mathbb{R}^{|V| \times d}$ be the embedding matrix over the base-token vocabulary, and for each modifier group $g$, let $E_g \in \mathbb{R}^{|M_g| \times d}$ be the corresponding modifier embedding matrix. The input embedding of \methodtoken position $i$ is
\[
\mathbf{e}(c_i) = E_{\text{base}}[b_i] + \sum_{g=1}^{G} E_g[m_i^{(g)}]
\]

In other words, each input position is the sum of a base-token embedding and a small set of modifier embeddings. Since the modifier vocabularies are small, this adds few new parameters relative to the lexical vocabulary.

On the output side, the model predicts the next \methodtoken in two stages: it first predicts the base token, and then predicts the modifier values conditioned on that base. Let $\mathbf{h}_i$ denote the final normalized hidden state at position $i$, and $U_{\text{base}} \in \mathbb{R}^{|V| \times d}$ the base-vocabulary unembedding matrix. The model first projects $\mathbf{h}_i$, yielding base-token logits:
\[
\boldsymbol{\ell}^{\text{base}}_i = U_{\text{base}} \mathbf{h}_i
\qquad
p(b_i \mid c_{<i}) = \mathrm{softmax}(\boldsymbol{\ell}^{\text{base}}_i)
\]
Let $\mathbf{u}_{b_i}$ denote the normalized unembedding vector of the selected base token. The modifier prediction head combines contextual and lexical representations as
\[
\boldsymbol{\ell}^{\mathrm{mod}}_i=W_h\mathbf{h}_i+\alpha_i W_b\mathbf{u}_{b_i},
\qquad \alpha_i=\sigma(\mathbf{w}_{\alpha}^{\top}\mathbf{h}_i)
\]
Here $W_h$ and $W_b$ project the hidden state and selected base-token representation into a shared modifier-logit space. The gate $\mathbf{w}_{\alpha}$ controls how strongly the selected base influences the modifier prediction. A refinement $W_r$ then lets the modifier logits influence one another:
\[
\widetilde{\boldsymbol{\ell}}^{\mathrm{mod}}_i=\boldsymbol{\ell}^{\mathrm{mod}}_i+W_r\operatorname{SiLU}(\boldsymbol{\ell}^{\mathrm{mod}}_i)
\]
Finally, the refined logits are partitioned into modifier groups, and a separate softmax is applied within each group, yielding the factorization\footnote{This factorization assumes conditional independence among modifier groups given the hidden state and base token. Jointly predicting common modifier bundles or autoregressively predicting modifier groups are natural extensions, which we defer to future work.}
\[
p(c_i \mid c_{<i})
=
p(b_i \mid c_{<i})
\prod_{g \in {G}}
p\!\left(m_i^{(g)} \mid b_i, c_{<i}\right)
\]
where ${G}$ denotes the modifier groups.\footnote{The modifier parameter overhead is small: the $\sim$100 English modifiers require 0.5M parameters for embeddings and prediction, under 0.1\% of the corresponding BPE model's parameters.}

During training, under teacher forcing, the modifier logits are conditioned on the gold base token. The loss therefore decomposes into a base-prediction term and a sum of modifier-prediction terms:
\[
\mathcal{L}
= - \sum_{i=1}^{N}
\left[
\log p(b_i \mid c_{<i})
+
\sum_{g \in {G}} \log p\!\left(m_i^{(g)} \mid b_i, c_{<i}\right)
\right]
\]
where inactive groups take their \emph{null} label. At inference time, the model first decodes a base token from $p(b_i \mid c_{<i})$ and then predicts the modifier values conditioned on that choice. The predicted \methodtoken is then deterministically rendered as text directly from the structured base-plus-modifier prediction, so generation remains standard at the level of text even though the model internally operates over a shorter sequence of structured units.

\paragraph{Evaluation.}
Because \methodbpe changes both the prediction unit and the factorization of text probability, its token-level likelihoods are not directly comparable to those of a standard autoregressive language model. A single \methodbpe prediction may jointly determine a base token along with a determiner, preposition, capitalization, and punctuation, whereas a standard BPE model would treat these as separate (and often more easily predictable) tokens. Moreover, each \methodbpe position scores every modifier group, including inactive labels for unused modifiers. Byte-normalized likelihood therefore remains well defined, but measures this full structured prediction problem rather than sequence compression or compute efficiency alone. We therefore evaluate \methodbpe primarily through downstream performance, including both likelihood-based benchmarks and open-ended generation tasks, providing a natural basis for comparison across different language modeling setups.

%% file: tables/tokenization_examples.tex
\begin{table}[t]
\centering
\scriptsize
\setlength{\tabcolsep}{4pt}
\begin{tabular}{>{\raggedright\arraybackslash}p{0.13\linewidth}
                >{\raggedright\arraybackslash}p{0.24\linewidth}
                >{\raggedright\arraybackslash}p{0.57\linewidth}}
\toprule
Text & BPE encoding \hfill \# tokens & \methodbpe\ encoding (1 \methodtoken; base = \texttt{ground}) \\
\midrule
\texttt{ground}
& \texttt{ground} \hfill 1
& --- \\

\texttt{the ground}
& \texttt{the | \_ground} \hfill 2
& \texttt{determiner=the} \\

\texttt{ The ground}
& \texttt{\_The | \_ground} \hfill 2
& \texttt{leading-space, determiner=the, capitalize-determiner} \\

\texttt{ The Ground}
& \texttt{\_The | \_Ground} \hfill 2
& \texttt{leading-space, capitalize, determiner=the, capitalize-determiner} \\

\texttt{In the ground}
& \texttt{In | \_the | \_ground} \hfill 3
& \texttt{preposition=in, capitalize-preposition, determiner=the} \\

\texttt{in the ground,}
& \texttt{in | \_the | \_ground | ,} \hfill 4
& \texttt{preposition=in, determiner=the, punct-suffix=,} \\

\texttt{(in the ground)}
& \texttt{( | in | the | ground | )} \hfill 5
& \texttt{punct-prefix=(, preposition=in, determiner=the, punct-suffix=)} \\
\bottomrule
\end{tabular}
\caption{\textbf{Examples of \methodbpe decompositions.}
A leading underscore in the BPE column denotes a space-prefixed token. Standard BPE assigns separate token identities to surface variants such as ``\emph{ground}'' and ``\spaceprefixed{\emph{Ground}}''. It also represents the examples above using 1--5 sequential tokens. \methodbpe instead keeps the lexical base token identity fixed and expresses variation through attached modifiers within a single structured \textsl{\methodtoken}.}
\label{tab:tokenization-examples}
\end{table}

%% file: figures/method_illustration.tex
\begin{figure}[t]
    \centering
    \includegraphics[width=0.98\linewidth]{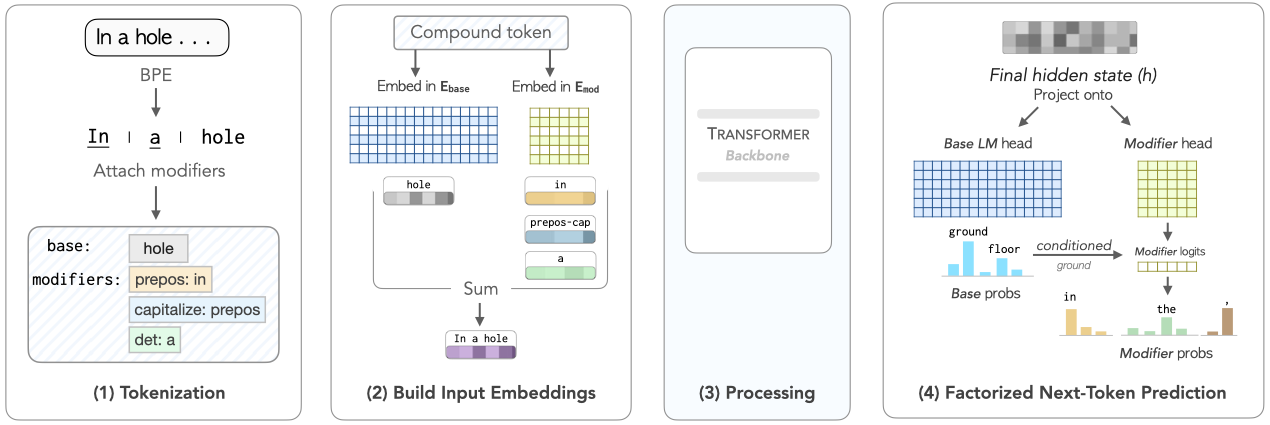}
    \caption{\textbf{Adapting language models to \methodbpe.} 
    (1) A span tokenized under \methodbpe is represented as a structured \textsl{\methodtoken} with a base token and attached modifiers. 
    (2) On the input side, the model embeds this unit compositionally by summing the base embedding and the embeddings of its active modifiers. 
    (3) The resulting sequence is processed by an otherwise standard transformer backbone. 
    (4) On the output side, next-token prediction is factorized: the model first predicts the next base token, and then predicts the modifier values conditioned on that base.}
    \label{fig:method-illustration}
\end{figure}

%% file: text/4.experiments.tex
\input{tables/main_results}

\section{Experiments}
\label{sec:experiments}
We test whether replacing BPE with \methodbpe improves downstream performance under controlled English pretraining from scratch~(\subsecref{main-results}), and whether its improved compression translates into better training efficiency~(\subsecref{speed-results}). The main English comparison holds the architecture, training recipe, token budget, and evaluation suite fixed while varying tokenization. We then analyze inference speed~(\subsecref{inference-speed}) and \methodbpe compression beyond English~(\subsecref{multilingual-potential}).

\subsection{Setup}
\label{subsec:experimental-setting}

We use ClimbMix to construct tokenizers and train models across all settings. Our main experiments use a standard decoder-only transformer architecture with a vocabulary size of 32k and untied input and output embeddings. We evaluate two model scales: a 24-layer model with 780M parameters and a 28-layer model with 1.3B parameters. Following~\cite{hoffmann2022training}, we train each model on 20 tokens per parameter, yielding approximately 15.6B and 26B tokens for the 24- and 28-layer models, respectively.\footnote{Notably, this comparison approximately matches training compute between BPE and \methodbpe models, but not raw-text exposure: \methodbpe encodes more text within the same number of tokens.} At the 24-layer scale, we also include SuperBPE~\citep{superbpe2025} as an additional baseline. Its transition point $t$ determines when tokenizer construction stops enforcing word boundaries and begins learning multiword tokens; we use $t=27\text{k}$.\footnote{We swept $t\in\{20\text{k},25\text{k},27\text{k}\}$ and selected by the highest 30-task average; the original paper used $t=180$k with a $200$k vocabulary, and pretrained much larger 8B models.} We run each experiment on 8 L40S GPUs. 
Evaluation uses SuperBPE's 30-task suite, spanning knowledge, reasoning, coding, reading, commonsense, language understanding, and string manipulation. See \appref{appendix-setup} for details on training, and~\appref{appendix-eval} for data and evaluation details.

We further experiment with the \nanochat ``GPT-2 speedrun'' benchmark~\citep{nanochat}, which measures the time required to train a model under the \nanochat harness to GPT-2-level capability~\citep{gpt2}, according to the DCLM CORE downstream evaluation suite~\citep{dclm}. We use the leading \nanochat submission at the time of writing---a 24-layer model with a custom architecture and optimization scheme~(see \appref{speedrun-details})---and evaluate BPE and \methodbpe under the same recipe, varying only tokenization, the corresponding input and output parameterization, and the training horizon~(\subsecref{modeling}).

\subsection{\methodbpe is more compact and more accurate than BPE}
\label{subsec:main-results}

Table~\ref{tab:main-results} reports the main matched training-compute comparison.\footnote{For readability, we omit 5 tasks on which all models score near zero, while still including them in the 30-task average; their results are given in~\appref{appendix-omitted}.} At both the 24-layer~(780M) and 28-layer~(1.3B) scales, \methodbpe outperforms standard BPE on the 30-task downstream suite while also reducing sequence length substantially. At the 24-layer scale, \methodbpe improves the 30-task average from 30.4 to 31.6 (+1.2), while SuperBPE matches the BPE baseline at 30.4. At the 28-layer scale, \methodbpe improves the average from 33.3 to 34.6~(+1.3).
Importantly, \methodbpe is also far more efficient than BPE. Across both scales, \methodbpe yields a 30\% reduction in sequence length, compared with 15\% for SuperBPE.

\paragraph{Tokenization-sensitive evaluation tasks.} Although \methodbpe improves the 30-task average, its largest regressions occur on LAMBADA and Dyck Languages. Examining their formats suggests that both are partly driven by evaluation inputs that are atypical under \methodbpe tokenization. LAMBADA asks the model to predict the final word withheld from a passage. If its prompt ends in an attachable marker, as in ``Tom opened the'' with target ``door'', withholding the target leaves ``the'' as a standalone base token, although \methodbpe would normally see it as a modifier. To test this boundary effect, we exclude prompts ending in an attachable determiner or preposition, retaining 2,660 of 5,000 examples. On this subset, the 24-layer \methodbpe model rises from 29.7 to 35.5 EM, whereas BPE changes from 34.5 to 32.8. Dyck Languages instead consists entirely of bracket strings such as ``(()())'', where punctuation itself is the content to be modeled. Because \methodbpe usually composes punctuation with lexical bases, it very rarely encounters such standalone punctuation tokens during pretraining. The original SuperBPE evaluation likewise reports regressions on both tasks~\citep{superbpe2025}. Together, these cases highlight that some benchmarks are sensitive to tokenizer design, requiring care when comparing tokenization methods.

\subsection{Nanochat speedrun}
\label{subsec:speed-results}

We next evaluate \methodbpe on the \nanochat speedrun benchmark~\citep{nanochat} which measures wall-clock training time needed to reach GPT-2-level capability, defined by a score of 25.65 on the CORE evaluation suite. The benchmark leaves architecture and training choices open. For a controlled tokenizer comparison, we use the latest \nanochat training recipe and configuration available at the time of writing, varying only the tokenization scheme, its corresponding input and output parameterization, and the training horizon.\footnote{See~\appref{speedrun-details} for the exact speedrun configuration and its adaptation to \methodbpe.}
Due to limited resources, we run our experiments on 8 L40S rather than the benchmark's standard H100. As such, these wall-clock times are not directly comparable to the public \nanochat leaderboard, but remain comparable across methods within our setting.

Table~\ref{tab:speedrun} shows that \methodbpe surpasses the BPE reference's performance with fewer training tokens and shorter wall-clock time.\footnote{For reference, training BPE for the smaller token budget used by \methodbpe reaches CORE 23.80.} Overall, this demonstrates that the sequence compression induced by \methodbpe translates into end-to-end training efficiency.

\input{tables/speedrun}

\subsection{Inference speed analysis}
\label{subsec:inference-speed}

To measure inference speed on identical generated text, we prompt the 28-layer vanilla models to copy the same CNN/DailyMail passages~\citep{cnn_dailymail} and report latency on 250 examples where both models reproduce the target exactly. As shown in~\tabref{inference-speed}, \methodbpe uses 32.5\% fewer decoding steps and reduces average generation time from 1.94 to 1.51 seconds despite its higher per-token cost. The resulting 1.28$\times$ speedup shows that sequence compression more than offsets the structured-prediction overhead.

%% file: tables/main_results.tex
\begin{table*}[!t]
\centering
\scriptsize
\setlength{\tabcolsep}{3pt}
\resizebox{0.75\textwidth}{!}{\begin{tabular}{llccccc}
\toprule
 &  & \multicolumn{3}{c}{24-layer (780M)} & \multicolumn{2}{c}{28-layer (1.3B)} \\
\cmidrule(lr){3-5}\cmidrule(lr){6-7}
Category & Task & BPE & SuperBPE\tiny{ ($t=27\text{k}$)} & \methodbpe & BPE & \methodbpe \\
\midrule
Knowledge & ARC-Easy \tiny{(MC)} & 68.3 & 69.7 & \textbf{74.1} & 73.8 & \textbf{76.1} \\
 & ARC-Challenge \tiny{(MC)} & 40.3 & 41.8 & \textbf{42.7} & 43.3 & \textbf{47.4} \\
 & Jeopardy \tiny{(EM)} & 5.9 & 5.7 & \textbf{12.0} & 11.8 & \textbf{18.0} \\
 & MMLU \tiny{(MC)} & 25.2 & \textbf{25.3} & 25.2 & 25.0 & \textbf{25.5} \\
 & OpenBookQA \tiny{(MC)} & 39.8 & \textbf{42.6} & 42.2 & \textbf{42.8} & 39.2 \\
 & TriviaQA \tiny{(EM)} & 9.1 & 9.2 & \textbf{14.6} & 14.3 & \textbf{18.7} \\
 & WikidataQA \tiny{(EM)} & 39.6 & 36.8 & \textbf{49.8} & 47.5 & \textbf{51.5} \\
\midrule
\multirow{2}{*}{\shortstack[l]{Reasoning \textbackslash \\ \;\; Symbolic}} & Arithmetic \tiny{(EM)} & 1.7 & \textbf{1.9} & 1.8 & \textbf{5.9} & 2.7 \\
 & LSAT-AR \tiny{(MC)} & \textbf{32.6} & 26.1 & 29.1 & 27.0 & \textbf{33.0} \\
 & Operators \tiny{(EM)} & 15.7 & 17.1 & \textbf{19.5} & \textbf{22.4} & 21.9 \\
\midrule
Reading & BoolQ \tiny{(MC)} & 56.0 & \textbf{62.4} & 61.9 & 63.5 & \textbf{67.6} \\
 & CoQA \tiny{(EM)} & 43.0 & 39.4 & \textbf{43.3} & 47.3 & \textbf{50.3} \\
 & DROP \tiny{(EM)} & 17.5 & 19.9 & \textbf{20.7} & 18.0 & \textbf{22.9} \\
 & SQuAD \tiny{(EM)} & 49.9 & 46.5 & \textbf{50.9} & 55.9 & \textbf{56.0} \\
\midrule
Commonsense & CommonsenseQA \tiny{(MC)} & \textbf{25.2} & 21.9 & 23.0 & 27.2 & \textbf{29.4} \\
 & COPA \tiny{(MC)} & 63.0 & 62.0 & \textbf{70.0} & 71.0 & \textbf{75.0} \\
 & PIQA \tiny{(MC)} & \textbf{74.6} & 73.8 & 74.5 & \textbf{76.7} & \textbf{76.7} \\
 & Winograd \tiny{(MC)} & 67.0 & \textbf{68.9} & 66.3 & 71.8 & \textbf{72.9} \\
 & Winogrande \tiny{(MC)} & \textbf{57.8} & 55.4 & 55.8 & 59.4 & \textbf{60.4} \\
\midrule
Language & HellaSwag \tiny{(MC)} & 56.7 & 56.4 & \textbf{57.5} & 62.9 & \textbf{63.7} \\
 & LAMBADA \tiny{(EM)} & 34.4 & \textbf{39.0} & 29.7 & \textbf{42.7} & 35.7 \\
 & Language ID \tiny{(EM)} & 24.4 & 24.3 & \textbf{24.9} & 24.3 & \textbf{25.2} \\
\midrule
String / formal & CS Algorithms \tiny{(EM)} & 39.4 & 39.4 & \textbf{40.7} & 39.5 & \textbf{45.8} \\
 & CUTE \tiny{(EM)} & 1.7 & 2.1 & \textbf{2.7} & 3.4 & \textbf{3.7} \\
 & Dyck Languages \tiny{(EM)} & \textbf{19.6} & 19.3 & 11.6 & \textbf{17.3} & 14.3 \\
\midrule
Aggregate & 30-task average & 30.4 & 30.4 & \textbf{31.6} & 33.3 & \textbf{34.6} \\
 & Relative sequence length & 1.00 & 0.85 & \textbf{0.70} & 1.00 & \textbf{0.70} \\
\bottomrule
\end{tabular}
}
\caption{\textbf{Main downstream results under matched training-compute budgets.} Models are trained for 15.6B tokens at the 24-layer, 780M scale and 26B tokens at the 28-layer, 1.3B scale. The primary comparison is BPE versus \methodbpe at each scale. At the 24-layer scale, we also include SuperBPE as a secondary baseline, using the best transition point from our sweep for a 32k vocabulary. Best results per task and model size are in \textbf{bold}.
}
\label{tab:main-results}
\end{table*}

%% file: tables/speedrun.tex
\begin{table*}[t]
    \centering
    \small
    \scriptsize
    \resizebox{0.65\textwidth}{!}{\begin{tabular}{lcccc}
        \toprule
        Model & CORE$\uparrow$ & Tokens to target $\downarrow$ & Hours to target $\downarrow$ & Speedup $\uparrow$ \\
        \midrule
        BPE & 25.91 & 5.84B & 7.92h & 1.00$\times$ \\
        \methodbpe & 26.00 & 4.74B & 6.45h & 1.23$\times$ \\
        \bottomrule
    \end{tabular}
    }
    \caption{\textbf{\nanochat speedrun results on 8 L40S GPUs.} We report CORE score at the stopping point, along with the training tokens and wall-clock hours used in each run. Times are comparable within this table, but not to the public \nanochat H100 leaderboard.}
    \label{tab:speedrun}
\end{table*}

%% file: text/5.multilingual_potential.tex
\subsection{Compression potential beyond English}
\label{subsec:multilingual-potential}
Finally, we test whether the same composition principle can shorten sequences beyond English in Spanish, German, Hindi, and Indonesian. For each language, we build 16k-entry BPE and \methodbpe tokenizers from the same data and vocabulary budget. We derive the \methodbpe modifier inventories automatically from UD annotations~\citep{de-marneffe-etal-2021-universal} using fixed extraction rules~(see~\appref{multilingual-details}), alongside shared spacing and punctuation cues, plus capitalization for the Latin-script languages. We measure sequence length directly, leaving multilingual pretraining and downstream evaluation for future work. As shown in~\tabref{multilingual-compression}, \methodbpe reduces sequence length by 18.0--31.8\% across the four languages, suggesting considerable compression opportunities beyond English.

\input{tables/inference_multilingual}

%% file: tables/inference_multilingual.tex
\begin{table*}[t]
    \centering
    \begin{minipage}[t]{0.48\textwidth}
        \centering
        \scriptsize
        \begin{tabular}{lcc}
            \toprule
            Metric & BPE & \methodbpe \\
            \midrule
            Gen. tokens $\downarrow$ & 120.7 & 81.5 \\
            Gen. time (s) $\downarrow$ & 1.94 & 1.51 \\
            ms / token & 16.1 & 18.6 \\
            Speedup $\uparrow$ & 1.00$\times$ & 1.28$\times$ \\
            \bottomrule
        \end{tabular}
        \captionof{table}{\textbf{Inference speed on text copying.} Decoding steps and wall-clock time for 28-layer models prompted to copy CNN/DailyMail passages.}
        \label{tab:inference-speed}
    \end{minipage}
    \hfill
    \begin{minipage}[t]{0.48\textwidth}
        \centering
        \scriptsize
        \begin{tabular}{lc}
            \toprule
            Language & Positions saved \\
            \midrule
            Spanish & 31.8\% \\
            German & 21.2\% \\
            Hindi & 20.8\% \\
            Indonesian & 18.0\% \\
            \bottomrule
        \end{tabular}
        \captionof{table}{\textbf{Compression beyond English.} Sequence-length reduction from \methodbpe relative to BPE with the same 16k vocabulary budget on held-out FineWeb-2 text.}
        \label{tab:multilingual-compression}
    \end{minipage}
\end{table*}

%% file: text/999.related_work.tex
\section{Related work}
\label{sec:related}

\methodbpe connects two lines of work: tokenization for compression, and compositional lexical representations. Standard BPE remains the dominant tokenizer in LMs~\citep{bpe2016sennrich}, with WordPiece~\citep{schuster-nakajima-2012-wordpiece,devlin-etal-2019-bert} and Unigram LM~\citep{kudo-2018-subword} serving as established alternatives. 
Recent work shows that tokenizer design affects both compression and downstream quality~\citep{bostrom-durrett-2020-byte,tokenizer-choice-2024,pickybpe2024}, while better compression alone does not guarantee better models~\citep{dagan2024getting,schmidt-etal-2024-tokenization}.
Our closest comparison is SuperBPE~\citep{superbpe2025}, which learns atomic multi-word surface tokens beyond pretokenization boundaries. \methodbpe instead represents frequent local markers as reusable modifiers, making the approaches potentially complementary: SuperBPE could learn phrases not already captured by modifier composition.
Like tokenization-free methods such as MEGABYTE and BLT, \methodbpe aims to cover more text per global modeling step~\citep{yu2023megabyte,pagnoni-etal-2025-byte}, but through explicit lexical composition rather than byte patches.

A second connection is to structured lexical representations.
Prior work has explored morphology-aware embeddings, tokenization, or pretraining objectives~\citep{matthews2018using,blevins2019better,gan2022morphte,cui2022lert,asgari2025morphbpe,lerner-yvon-2025-unlike}. 
Our input-output parameterization follows the modeling scheme introduced by \citet{vocab_diet}, but targets grammatical markers rather than morphology. 
Related NMT systems likewise separate lexical content from surface or linguistic attributes, using inline casing tags or input features~\citep{berard-etal-2019-naver,etchegoyhen-gete-2020-case} and factored prediction of lemmas with POS, gender, and number~\citep{garcia-martinez-etal-2016-factored}. \methodbpe extends this separation to the tokenization and language-modeling interface.

%% file: text/999.conclusion.tex
\section{Conclusion}
\label{sec:conclusion}

We introduced \methodbpe, a compositional tokenization scheme that represents text as lexical base tokens with attached surface modifiers. Across two model scales, \methodbpe shortens sequences by 30\% and improves average downstream performance and end-to-end training efficiency.
More broadly, these results show that structure conventionally spread across token positions can instead be modeled within each prediction, making token representation part of the language-modeling interface rather than merely a preprocessing choice.
This opens a broad and underexplored design space between surface-form tokenization and fully learned latent interfaces. By allowing each sequence position to carry more information, structured tokenization offers a promising foundation for future language models.

%% file: text/999.acknowledgments.tex
\section*{Acknowledgments}

This work was supported in part by the Israel Science Foundation (grant no.\ 2045/21) and by NSF-BSF grant 2020793. 
We also thank NVIDIA Corporation for support through the NVIDIA Academic Grant Program, which provided access to A100 GPUs used for preliminary experiments.

%% file: text/999.appendix.tex
\section{Additional results}
\label{app:appendix-analysis}

\subsection{Grammatical markers as standalone BPE tokens}
\label{app:appendix-markers-breakdown}
Figure~\ref{fig:grammar-marker-detail} breaks down the aggregate categories in Figure~\ref{fig:grammar-marker-share}, showing which individual determiners, prepositions, and punctuation marks account for most of the mass discussed in Section~\ref{sec:compression-motivation}.

\input{figures/grammar_marker_detail_appendix}

\subsection{Additional downstream results}
\label{app:appendix-omitted}
Table~\ref{tab:appendix-omitted-results} reports the five tasks omitted from Table~\ref{tab:main-results} for readability. These tasks are still included in the 30-task average reported in the main text.

\section{\methodbpe implementation details}
\label{app:tokenizer-details}

This section records additional details on the \methodbpe tokenizer construction. We specify the closed modifier inventories used in the reported experiments and the deterministic, second-stage composition pass.

\subsection{Modifier inventories}

All reported \methodbpe experiments use the same English modifier inventory: determiner and preposition groups, prefix- and suffix-punctuation groups, a leading-space indicator, and capitalization indicators for the base token, determiner, and preposition. The determiner and preposition groups contain ten and nine active entries, respectively. Tables~\ref{tab:appendix-modifier-inventory} and~\ref{tab:appendix-punctuation-inventory} list the exact supported values. 

The modifier inventory is deliberately small and restricted to markers that can be identified and reconstructed from their surface form and local token order. No syntactic parsing is used. Listed forms are attached when they occur in an eligible local configuration, without determining their grammatical role. We therefore exclude forms such as ``before'' and ``past'', pronouns and auxiliaries that require role disambiguation, and forms written in ALLCAPS or camelCase. When a form is excluded or cannot be attached, the tokenizer leaves it as an ordinary BPE token in the final representation.

\paragraph{Modifier usage.} In FineWeb text, 85.9\% of \methodbpe token positions carry at least one modifier, 13.4\% are base-only lexical positions, and 0.7\% are standalone inventory markers that could not be attached. Modifier-bearing positions are dominated by the leading-space modifier, which factors surface variation out of the vocabulary but does not itself absorb a separate BPE position.

\subsection{\methodbpe composition pass}

At inference time, we apply a deterministic left-to-right composition pass to the first-stage BPE stream. The pass maintains a pending modifier state, absorbs eligible prefix punctuation, determiners, and prepositions before a lexical base, and attaches suffix punctuation after the lexical span is identified. Leading-space and capitalization cues are attached to the first token of a multi-token base span, while suffix punctuation is attached to its final token. Tokens that do not participate in a valid local composition remain ordinary base tokens. Algorithm~\ref{alg:cobpe-inference} gives the exact pseudocode for the current implementation.

\input{tables/appendix_omitted_results}

\begin{algorithm}[t]
\caption{\methodbpe tokenization at inference time}
\label{alg:cobpe-inference}
\begin{algorithmic}[1]
\Require text $x$, base tokenizer $\mathrm{BPE}$, modifier inventories
\Ensure base-token ids $Y$ and modifier rows $M$

\State $R \gets \mathrm{BPE.encode}(x)$
\State $Y \gets [\,]$, $M \gets [\,]$
\State $p \gets \mathrm{EmptyModifier}()$
\State $i \gets 0$

\While{$i < |R|$}
    \If{$R_i$ is attachable suffix punctuation for the previous emitted token}
        \State attach it to the last modifier row in $M$
        \State $i \gets i+1$; \textbf{continue}
    \EndIf

    \If{$R_i$ is attachable prefix punctuation, determiner, or preposition}
        \State record it in pending modifier row $p$
        \State $i \gets i+1$; \textbf{continue}
    \EndIf

    \If{a lexical base span starts at position $i$}
        \State merge $p$ with the span's own space/capitalization information
        \State place span-initial modifiers on the first token and suffix punctuation on the last
        \State append the span to $(Y, M)$
        \State clear $p$
        \State advance $i$ past the span
        \State \textbf{continue}
    \EndIf

    \State emit $R_i$ literally, merging in any pending modifiers
    \State clear $p$
    \State $i \gets i+1$
\EndWhile

\If{pending modifiers remain}
    \State emit them literally
\EndIf

\State \Return $(Y, M)$
\end{algorithmic}
\end{algorithm}

\input{tables/appendix_modifier_inventory}
\input{tables/appendix_punctuation_inventory}

\subsection{Optimizing \methodbpe tokenization speed}

The reported tokenizer is a deliberately simple two-stage research implementation rather than a latency-optimized one. In principle, the same composition logic could be fused directly into BPE tokenization, and should bring \methodbpe encoding speed closer to that of regular BPE. For research and experimentation purposes, we chose to use an explicit second pass for the reported experiments, as it makes the modifier inventory and attachment rules transparent and easy to vary.

\section{Experimental setup}
\label{app:appendix-setup}

\subsection{Data and tokenizer training}

All experiments use ClimbMix data for constructing tokenizers and training models. We keep the training corpus, validation split, and tokenizer-training sample fixed across tokenization variants. All tokenizers use a vocabulary size of 32k.

Tokenizer training follows the dedicated tokenization scripts in the nanochat codebase. The baseline BPE tokenizer is trained on 2B characters from the ClimbMix corpus, with each document truncated to at most 10k characters and random seed 42. The BPE tokenizer trained on space- and capitalization-normalized text that underlies \methodbpe uses the same 2B-character sample with the same document cap and seed.

\subsection{Model configurations}

All main models are decoder-only transformers with untied input and output embeddings. Table~\ref{tab:appendix-model-configs} gives the exact architecture settings for the 24-layer and 28-layer runs.

\input{tables/appendix_model_configs}

\subsection{Modifier prediction head architecture}
\label{app:modifier-head}

We use the base-conditioned gated head described in~\subsecref{modeling}. Let $M$ be the total number of modifier classes. The head combines a linear projection of the model's final hidden state with a gated projection of the selected base-token representation:
\[
\boldsymbol{\ell}^{\mathrm{mod}}_i = W_h\mathbf{h}_i + \alpha_i W_b\mathbf{u}_{b_i},
\qquad
\alpha_i=\sigma(\mathbf{w}_{\alpha}^{\top}\mathbf{h}_i)
\]
The head then applies a refinement layer that adds a learned nonlinear transformation of the full logit vector:
\[
\widetilde{\boldsymbol{\ell}}^{\mathrm{mod}}_i
=\boldsymbol{\ell}^{\mathrm{mod}}_i+W_r\operatorname{SiLU}(\boldsymbol{\ell}^{\mathrm{mod}}_i).
\]
The refined logits are split into modifier groups. In our inventory, $M=124$. During teacher forcing, the head receives the gold base token, while during generation it receives the selected base token. The gate is initialized at equal weighting, $W_r$ is initialized to zero, and the modifier-group losses are summed with unit weight.

\subsection{Optimization}

Across the main matched training-compute runs, we keep the architecture preset, context length, optimizer family~\citep[Muon;][]{muon}, learning-rate schedule, checkpoint cadence, and data source fixed while varying only the tokenization scheme and, for \methodbpe, the corresponding structured input and output parameterization. The main runs use a vanilla variant of the \nanochat codebase and architecture preset, stripped of the speedrun-specific architecture and optimization tricks (e.g., no value embeddings or residual connections across non-contiguous layers), with a maximum sequence length of 2048, a target parameter-to-data ratio of 20, and the DeepSeek-style 80/10/10 learning-rate step-based schedule~\citep{deepseek_llm}. The reported speedrun comparison does not include midtraining or SFT. It evaluates the pretraining-to-CORE target portion of the \nanochat recipe.

\section{Evaluation details}
\label{app:appendix-eval}

We follow SuperBPE's prompting and scoring protocol for the evaluation tasks. Unless noted otherwise, each task uses five in-context examples. Multiple-choice tasks are scored by accuracy over the answer choices, open-ended tasks are scored by exact match against the target after task-specific normalization of case, punctuation, and articles, following the SQuAD protocol, and coding tasks are scored with pass@10. We evaluate up to 5,000 examples per task, or the full split when the dataset is smaller.

\subsection{Knowledge}

The knowledge group contains ARC-Easy and ARC-Challenge, which are four-way science multiple-choice benchmarks drawn from grades 3--9 science exams \citep{clark-etal-2018-think}; Jeopardy, an open-ended quiz benchmark built from ``Jeopardy!'' questions,\footnote{\url{https://www.kaggle.com/datasets/tunguz/200000-jeopardy-questions}} TriviaQA, an open-ended factual QA task \citep{joshi-etal-2017-triviaqa}; WikidataQA, a factual continuation task; MMLU, which covers broad academic and professional subject areas \citep{hendrycks-etal-2021-measuring}; and OpenBookQA, a science QA benchmark requiring fact retrieval and light reasoning \citep{mihaylov-etal-2018-suit}.

\subsection{Reasoning and symbolic tasks}

This group includes Arithmetic on the \texttt{2da}, \texttt{2dm}, and \texttt{2ds} subsets \citep{brown-etal-2020-language}; LSAT-AR for analytical-reasoning multiple choice \citep{zhong-etal-2022-analytical, zhong-etal-2024-agieval}; BIG-Bench Operators; Repeat-Copy-Logic \citep{austin-etal-2021-program}; and GSM8K \citep{cobbe-etal-2021-training}. For GSM8K, the demonstrations include worked solutions with calculator annotations removed, and the model is prompted to produce the final numeric answer after the delimiter \texttt{\#\#\#\#}.

\subsection{Reading}

The reading-comprehension group contains BoolQ \citep{clark-etal-2019-boolq}, CoQA \cite{reddy-etal-2019-coqa}, DROP \citep{dua-etal-2019-drop}, HotpotQA \citep{yang-etal-2018-hotpotqa}, and SQuAD \citep{rajpurkar-etal-2016-squad}. CoQA is evaluated without additional in-context examples because the conversation history already provides the effective demonstrations; the remaining tasks use the standard five-shot protocol.

\subsection{Commonsense}

The commonsense group contains CommonsenseQA \citep{talmor-etal-2019-commonsenseqa}, COPA \citep{roemelle-etal-2011-choice, kavumba-etal-2019-choosing}, PIQA \citep{bisk-etal-2020-piqa}, Winograd \citep{levesque-etal-2012-winograd}, and Winogrande \citep{sakaguchi-etal-2021-winogrande}, all scored by exact match over the provided answer options.

\subsection{Language understanding}

This group contains HellaSwag \citep{zellers-etal-2019-hellaswag}, LAMBADA \citep{paperno-etal-2016-lambada}, and BIG-Bench Language Identification. HellaSwag is four-way multiple choice, while LAMBADA and Language Identification are open-ended exact-match tasks.

\subsection{String and formal tasks}

The string/formal group contains BIG-Bench CS Algorithms \citep{srivastava-etal-2023-beyond}, CUTE \citep{edman-etal-2024-cute}, and BIG-Bench Dyck Languages, all scored by exact match on the generated answer string.

\subsection{Coding tasks}

The coding tasks are HumanEval \citep{chen-etal-2021-evaluating} and MBPP~\citep{austin-etal-2021-program}, both evaluated zero-shot, following prior work. For HumanEval, we use the stop tokens \verb|\nclass|, \verb|\ndef|, \verb|\n#|, and \verb|\nif|; sample 20 continuations with top-$p=0.95$ and temperature $0.8$; and cap generation at 128 new tokens. MBPP uses the same decoding and functional-correctness setup. We report pass@10, estimated from the 20 sampled solutions per problem.

\section{Speedrun details}
\label{app:speedrun-details}

We use the latest \nanochat GPT-2 training recipe and speedrun configuration available as of July 24, 2026. We run both methods on 8 L40S GPUs rather than the benchmark's standard H100 hardware and keep the backbone and per-step optimization configuration fixed. We vary the tokenization, the corresponding \methodbpe input and output parameterization, and the training horizon between methods.

Table~\ref{tab:appendix-speedrun-setup} gives the exact 24-layer configuration. Relative to the H100 recipe, the L40S runs disable FP8 training and use the full-window attention pattern \texttt{L} rather than the short-window \texttt{SSSL} pattern. The reported \methodbpe speedrun uses the same base determiner profile and existing preposition profile as the main experiments.

\subsection{\nanochat-specific residual mechanisms}
\label{app:speedrun-mechanisms}

The speedrun recipe uses smear, which mixes a gated amount of the previous input embedding into the current one, and backout, which subtracts a learned fraction of a middle-layer representation before next-token prediction.

For \methodbpe, both mechanisms operate only on the lexical base stream. Smear precedes the addition of modifier embeddings, backout affects only the base-token prediction, and modifier prediction uses the pre-backout transformer representation. Modifier embeddings therefore participate in the shared transformer computation without forming a separate residual stream.

\input{tables/appendix_speedrun_setup}

\section{Multilingual tokenization analysis}
\label{app:multilingual-details}

This appendix details the multilingual tokenizer analysis summarized in~\subsecref{multilingual-potential}. We compare matched BPE and \methodbpe tokenizers, describe the automatically extracted modifier inventories, and document the validation procedure. The full inventories reduce positions by 18.0--31.8\% across the four languages.

\subsection{Data and construction}

We analyze Spanish, German, Hindi, and Indonesian on FineWeb-2~\citep{fineweb2}. For each language, we build matched 16k-entry BPE and \methodbpe tokenizers from the same training text and vocabulary budget, then measure sequence length on held-out text. All languages use the same whitespace-delimited lexical bases, spacing modifiers, and punctuation modifiers, with capitalization modifiers for the Latin-script languages. The training split is capped at 200,000 documents, with at most 10,000 Unicode characters per document. Evaluation uses 10,000 documents from the disjoint test split, capped at 10,000 characters per document. We report tokenizer compression only.

\subsection{Automatic modifier extraction}

We derive language-specific determiner and adposition inventories automatically from Universal Dependencies v2.18 annotations. The extraction retains frequent forms with consistent annotations and a consistent position relative to the lexical base, producing closed lists of surface strings that can be applied with exact matching and local attachment rules, without a language-specific parser. Forms that do not meet the filters remain ordinary BPE tokens. Hindi additionally uses a Devanagari-aware split and longest-match rule for compound postpositions, in which a multiword expression such as \textit{ke liye} (``for'') functions as a single postposition. Capitalization modifiers are used for Spanish, German, and Indonesian.

The inventories are conservative rather than exhaustive. For Indonesian, they contain four pre-base determiners and nine pre-base adpositions. Because the shared modifier interface provides one determiner group for pre-base markers, post-base determiners and multiword or affixed components remain ordinary BPE tokens.

\begin{table}[H]
\centering
\small
\setlength{\tabcolsep}{5pt}
\begin{tabular}{lccc}
\toprule
Language & Determiners & Adpositions & Attachment direction \\
\midrule
Spanish & 17 & 14 & before \\
German  & 24 & 16 & before \\
Hindi   & 5 & 19 + 15 compound & before / after \\
Indonesian & 4 & 9 & before \\
\bottomrule
\end{tabular}
\caption{\textbf{Language-specific modifier inventories.} Counts give the retained surface forms after frequency, annotation-consistency, and direction filtering. Attachment indicates whether the markers normally occur before or after the lexical base. Hindi reports ordinary adpositions plus compound postpositions.}
\label{tab:multilingual-inventories}
\end{table}

\section{Additional related work}
\label{app:appendix-related-work}

Other recent work aims to make each autoregressive step carry more future information, but without changing tokenization. Hidden-state analyses show that representations at one position already contain substantial signal about subsequent tokens \citep{pal-etal-2023-future,wu2024do,samragh2025your}. Multi-token prediction exploits this by training models to predict multiple future tokens from a shared trunk \citep{gloeckle2024better}, while related inference-time approaches propose or decode multiple future tokens or characters and then verify them \citep{cai2024medusa,ankner2024hydra,ben2025spellm,yan-etal-2025-decoding}. These methods are complementary to ours: they modify the objective or decoding procedure, whereas \methodbpe changes the representational unit. Analyses of depth also suggest that easy, high-frequency items are resolved earlier in the network, with function words and punctuation predicted earlier on average than nouns and verbs \citep{gupta2026how}, consistent with our motivation for modeling such markers compositionally.

%% file: figures/grammar_marker_detail_appendix.tex
\begin{figure}[h]
    \centering
    \includegraphics[width=\linewidth]{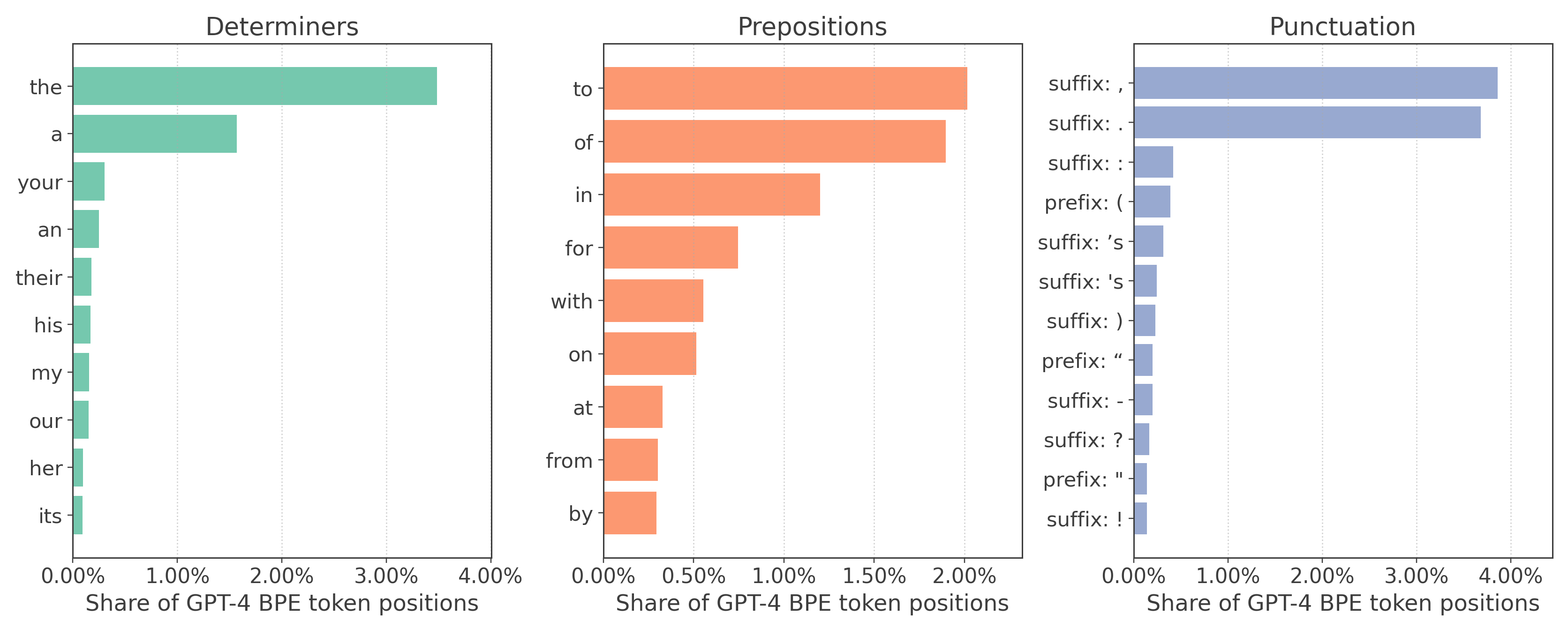}
    \caption{\textbf{Detailed breakdown of frequent grammatical markers.} Top marker-level shares within determiners, prepositions, and punctuation under GPT-4 tokenization on FineWeb text.}
    \label{fig:grammar-marker-detail}
\end{figure}

%% file: tables/appendix_omitted_results.tex
\begin{table*}[t]
\centering
\scriptsize
\setlength{\tabcolsep}{4pt}
\resizebox{0.78\textwidth}{!}{\begin{tabular}{llccccc}
\toprule
 &  & \multicolumn{3}{c}{24-layer (780M)} & \multicolumn{2}{c}{28-layer (1.3B)} \\
\cmidrule(lr){3-5}\cmidrule(lr){6-7}
Category & Task & BPE & SuperBPE\tiny{ ($t=27\text{k}$)} & \methodbpe & BPE & \methodbpe \\
\midrule
Reasoning / symbolic & Repeat-Copy-Logic \tiny{(EM)} & 0.0 & \textbf{3.1} & 0.0 & 0.0 & \textbf{3.1} \\
Reasoning / symbolic & GSM8K \tiny{(EM)} & \textbf{2.7} & 2.4 & 2.0 & \textbf{2.7} & 1.3 \\
Reading & HotpotQA \tiny{(EM)} & 0.0 & 0.0 & 0.0 & \textbf{0.0} & \textbf{0.0} \\
Coding & HumanEval \tiny{(P@10)} & 0.0 & 0.0 & 0.0 & \textbf{0.3} & 0.0 \\
Coding & MBPP \tiny{(P@10)} & 0.0 & 0.0 & 0.0 & \textbf{0.0} & \textbf{0.0} \\
\bottomrule
\end{tabular}}
\caption{\textbf{Results for the five tasks omitted from Table~\ref{tab:main-results}.} These tasks are included in the 30-task average reported in the main text.}
\label{tab:appendix-omitted-results}
\end{table*}

%% file: tables/appendix_modifier_inventory.tex
\begin{table*}[t]
\centering
\small
\setlength{\tabcolsep}{6pt}
\begin{tabular}{p{0.23\textwidth}p{0.68\textwidth}}
\toprule
Group & Values used in the reported experiments \\
\midrule
Leading space & Binary indicator: absent, present. \\
Preposition & No modifier, \texttt{by}, \texttt{at}, \texttt{of}, \texttt{to}, \texttt{in}, \texttt{on}, \texttt{with}, \texttt{for}, \texttt{from}. \\
Determiner & No modifier, \texttt{the}, \texttt{a}, \texttt{an}, \texttt{my}, \texttt{your}, \texttt{his}, \texttt{her}, \texttt{our}, \texttt{their}, \texttt{its}. \\
Preposition capitalization & Binary indicator: lowercase preposition, titlecased preposition. \\
Determiner capitalization & Binary indicator: lowercase determiner, titlecased determiner. \\
Base capitalization & Binary indicator: lowercase base, titlecased base. \\
\bottomrule
\end{tabular}
\caption{\textbf{Non-punctuation modifier inventory for the reported \methodbpe runs.} Each group includes a no-modifier state used when that modifier is inactive.}
\label{tab:appendix-modifier-inventory}
\end{table*}

%% file: tables/appendix_punctuation_inventory.tex
\begin{table*}[t]
\centering
\small
\setlength{\tabcolsep}{6pt}
\begin{tabular}{p{0.23\textwidth}p{0.68\textwidth}}
\toprule
Group & Values used in the reported experiments \\
\midrule
Prefix punctuation & No modifier, \texttt{"}, \texttt{'}, \texttt{`}, \texttt{(}, \texttt{[}, \{, \texttt{-}, \texttt{---}, \texttt{‘}, \texttt{’}, \texttt{“}, \texttt{”} \\
Suffix punctuation & No modifier, \texttt{"}, \texttt{'}, \texttt{`}, \texttt{)}, \texttt{]}, \}, \texttt{.}, \texttt{!}, \texttt{?}, \texttt{,}, \texttt{;}, \texttt{:}, \texttt{-}, \texttt{---}, \texttt{’}, possessive endings \texttt{'s}, \texttt{s'}, \texttt{’s}, \texttt{s’} \\
\bottomrule
\end{tabular}
\caption{\textbf{Exact punctuation strings supported by the \methodbpe inventory.} These are closed lists; punctuation outside them remains in the base-token stream.}
\label{tab:appendix-punctuation-inventory}
\end{table*}

%% file: tables/appendix_model_configs.tex
\begin{table}[t]
\centering
\small
\begin{tabular}{lcc}
\toprule
 & 24 layers & 28 layers \\
\midrule
Parameters & 780M & 1.3B \\
Layers & 24 & 28 \\
Hidden size & 1536 & 2048 \\
Intermediate size & 6144 & 6144 \\
Head dimension & 128 & 128 \\
Query heads & 12 & 16 \\
KV heads & 12 & 16 \\
Training tokens & 15.6B & 26B \\
\bottomrule
\end{tabular}
\caption{\textbf{Exact model configurations for the main experiments.}}
\label{tab:appendix-model-configs}
\end{table}

%% file: tables/appendix_speedrun_setup.tex
\begin{table*}[t]
\centering
\small
\setlength{\tabcolsep}{6pt}
\begin{tabular}{p{0.28\textwidth}p{0.60\textwidth}}
\toprule
Setting & Value \\
\midrule
Backbone scale & 24 layers, hidden size 1536, intermediate size 6144, head dimension 128. \\
Attention heads & 12 query heads and 12 KV heads. \\
Vocabulary / context & 32k vocabulary, maximum sequence length 2048. \\
Hardware & 8$\times$ L40S GPUs. \\
Batching & Device batch size 16; total batch size 1.05M tokens. \\
Target param:data ratio & 8.0 for BPE; 6.5 for the \methodbpe variant \\
Architecture details & Residual scalars, value embeddings on alternating layers, QK-norm scale 1.15, and MLP input-projection init scale 0.5. Both BPE and \methodbpe use smear, which mixes the previous input embedding into the current one, and backout, which removes a learned middle-layer component before prediction. For \methodbpe, these operate only on the base stream. \\
Optimization details & 40 steps of warmup, Muon final momentum of 0.97, and LR linear decay in the last 65\% steps \\
Attention pattern on L40S & Full-window pattern \texttt{L}. Relative to the H100 recipe, the L40S runs do not use FP8 and do not use the short-window \texttt{SSSL} pattern. \\
Data input & ClimbMix, tokenized on the fly. \\
\bottomrule
\end{tabular}
\caption{\textbf{Exact configuration for the reported \nanochat speedrun comparison.} Both methods share the backbone and per-step optimization configuration. The target parameter-to-data ratio sets the training horizon for each method.}
\label{tab:appendix-speedrun-setup}
\end{table*}